\documentclass[final]{l4dc2024}
\makeatletter
\let\Ginclude@graphics\@org@Ginclude@graphics
\makeatother
\usepackage{times}

\DeclareMathOperator{\trace}{tr}
\DeclareMathOperator{\dist}{dist}
\DeclareMathOperator{\Proj}{Proj}
\DeclareMathOperator{\Exp}{Exp}
\DeclareMathOperator{\grad}{grad}
\DeclareMathOperator{\Hess}{Hess}
\newcommand{\calM}{\mathcal{M}}
\newcommand{\calL}{\mathcal{L}}
\newcommand{\calT}{\mathcal{T}}

\newtheorem{assum}[theorem]{Assumption}

\title[Automatic Spherical Gradient Descent]{Automatic Optimisation of Normalised Neural Networks}

\author{\Name{Namhoon Cho} \Email{n.cho@cranfield.ac.uk}
 \AND
 \Name{Hyo-Sang Shin} \Email{h.shin@cranfield.ac.uk}\\
    \addr Centre for Autonomous and Cyber-Physical Systems, Cranfield University, United Kingdom
}

\begin{document}

\maketitle

\begin{abstract}%
We propose automatic optimisation methods considering the geometry of matrix manifold for the normalised parameters of neural networks. Layerwise weight normalisation with respect to Frobenius norm is utilised to bound the Lipschitz constant and to enhance gradient reliability so that the trained networks are suitable for control applications. Our approach first initialises the network and normalises the data with respect to the $\ell^{2}$-$\ell^{2}$ gain of the initialised network. % The issue in update steps is that widely-used first-order optimisation algorithms assuming Euclidean parameter space are inconsistent with the manifold geometry of normalised parameters. 
Then, the proposed algorithms take the update structure based on the exponential map on high-dimensional spheres. Given an update direction such as that of the negative Riemannian gradient, we propose two different ways to determine the stepsize for descent. The first algorithm utilises automatic differentiation of the objective function along the update curve defined on the combined manifold of spheres. The directional second-order derivative information can be utilised without requiring explicit construction of the Hessian. The second algorithm utilises the majorisation-minimisation framework via architecture-aware majorisation for neural networks. With these new developments, the proposed methods avoid manual tuning and scheduling of the learning rate, thus providing an automated pipeline for optimizing normalised neural networks.
\end{abstract}


\section{Introduction}
Boundedness of the Lipschitz constant of a neural network is particularly important in control applications. The existence of unique solution and the stability properties of a nonlinear dynamics represented by Ordinary Differential Equations (ODEs) requires Lipschitz continuity of the ODE function. In the context of residual dynamics learning, the Lipschitz constant of the learned model determines the ultimate bound and convergence rate for the tracking error \cite{Shi_2019}. For these reasons, it is desired to bound or specify the Lipschitz constant of the neural network embedded in a control algorithm with a known value. Multiple approaches have been developed to certifiably bound the Lipschitz constant of neural networks; direct parameterisation guaranteeing bounded Lipschitz constant \cite{Wang_2023}, LipNet architecture which is a universal Lipschitz function approximator \cite{Anil_2019, Zhou_2022}, and spectral normalisation \cite{Miyato_2018, Lin_2021, Bartlett_2017}.

Layerwise normalisation has been adopted to improve training stability and generalisation. Serial composition of normalised weight matrices and Lipschitz continuous activation layers lead to the bounded Lipschitz constant for the entire neural network. Spectral normalisation is known as a means to keep the gradients within a reliable range \cite{Miyato_2018, Lin_2021}. Also, spectral normalisation of deep neural networks has demonstrated its effectiveness in achieving good generalisation characteristics \cite{Bartlett_2017, Liu_2020}. Notably, spectral normalisation turns out to be central to achieve reliable learning-based control as shown in \cite{Shi_2019, Shi_2022, Shi_2023, OConnell_2022}. 

The problem is that widely-used gradient-descent-based optimisation algorithms assuming Euclidean geometry of the parameter space are inconsistent with layerwise normalisation. In principle, the Riemannian-gradient-based optimisation on certain matrix manifolds can be performed to account for the curved space of normalised parameters \cite{Absil_2008, Boumal_2023}. In addition, taking the parameter geometry into account through the choice of a right distance function as i) the regularisation function or as ii) the trust-region penalty function added to the local objective surface often leads to more efficient optimisation. However, the Riemannian gradient descent methods have not seen widespread use as compared to the projected variant of the Euclidean gradient descent methods. Also, computationally expensive Riemannian Newton method is not suitable for deep learning where light computation per iteration is desired for scalability.

The attempts to develop optimisation algorithms considering the layerwise norm of parameters and the product structure of neural networks led to the architecture-aware optimisation methods studied in \cite{Bernstein_2020, Liu_2021, Bernstein_2023a, Bernstein_2023b}. These studies investigated the perturbation bounds \cite{Bernstein_2020} and derived optimisation algorithms correspondingly. As a result, the architecture-aware optimisation methods substantially reduced the burden to perform expensive search of the learning rate over a log-scale grid, leading to more automated learning pipelines along with the Automatic Differentiation (AD) tools. % Empirical results showed consistent learning performance over a range of tasks and network architectures, e.g., widths and depths, without learning rate tuning \cite{Bernstein_2023a}.

This study aims to enable automated optimisation of the neural networks with layerwise normalisation with respect to the Frobenius norm by explicitly considering the motion of parameters on spherical manifolds. The necessity to bound the Lipschitz constant of neural networks is considered in the previous works on architecture-aware optimisation through initialisation \cite{Bernstein_2023a} and explicit application of an appropriate prefactor to scale the update \cite{Bernstein_2020}. Nonetheless, the update process of those prior works does not strictly capture the structure of a curve on sphere. Also, the hyperparameter-free Majorisation-Minimisation (MM) optimisation method developed in \cite{Streeter_2023a} based on automatic bounding of the Taylor remainder series \cite{Streeter_2023b} does not consider the curved geometry of the parameter space. Optimisation dynamics for neural networks with normalisation layers was studied in \cite{Wan_2021, Roburin_2022b} from a geometric perspective. Although these works addressed optimisation of normalised neural networks as the motion on high-dimensional spheres, they focused more on analysing the effective learning dynamics characteristics of existing optimisation methods such as stochastic gradient decent and Adam rather than on developing tailored optimisation algorithms.

This study presents optimisation methods for layerwise-normalised neural networks. The preprocessing step is to properly scale the data considering the $\ell^{2}$-$\ell^{2}$ gain of the randomly initialised network. Given the structure of spherical motion for the update, the approach first specifies the update direction and then determines the stepsize at each iteration. The proposed methods differ in how the stepsize is determined. The first method utilises the information of second-order directional derivative along the chosen update direction. This method is Hessian-free in the sense that it requires neither the explicitly-constructed Hessian matrix needed in the Newton methods nor the Hessian-vector product \cite{Pearlmutter_1994} needed in the conjugate-gradient-based methods \cite{Martens_2010}. % The only additional information required as compared to the vanilla Euclidean gradient descent method is the second-order directional derivative of the objective function with respect to the scalar variable that parameterises the update curve on the unit-norm sphere of matrices. 
Another method utilises the MM framework for which the minimisation step determines the stepsize that minimises the majorant. % Sinusoidal majorisation strategies are developed depending on the simplifying assumptions that relate functional perturbation to parameter perturbation.

\section{Preliminaries}
\subsection{Notation}
In this study, $\left\| \cdot \right\|$ denotes Frobenius norm for matrices and $\ell^{2}$-norm for vectors. The $\ell^{2}-\ell^{2}$ operator norm, i.e., maximum singular value, of matrices will be denoted by $\left\| \cdot \right\|_{op}$.

\subsection{Geometry of Matrix Sphere}
The sphere $\mathbb{S}_{\mu}^{nm-1}$ is a set of $n \times m$ real matrices with constant Frobenius norm $\mu$, i.e., $\mathbb{S}_{\mu}^{nm-1}:=\left\{\left. X\in\mathbb{R}^{n \times m} \right| \left\|X\right\|=\mu\right\}$ where $\left\| \cdot\right\|$ refers to the norm in the ambient space. It can be defined as a Riemannian submanifold $\calM:=\mathbb{S}_{\mu}^{nm-1}$ of the ambient Euclidean space $\mathbb{R}^{n \times m}$ equipped with the Frobenius inner product given by $\left<U,V\right>:=\trace\left(U^{T}V\right)$ where $\left<\cdot,\cdot\right>$ denotes the inner product in the ambient space which satisfies $\left\|X\right\| = \sqrt{\left<X,X\right>}$. The tangent space at $X$ is $\calT_{X}\mathcal{M} = \left\{\left.U\in \mathbb{R}^{n \times m}\right|\left<X,U\right>=0\right\}$. A distance function is given by $\dist\left(X,Y\right) = \cos^{-1}\left(\left<X,Y\right> / \mu^2\right)$ for $X,Y \in \mathcal{M}$. The tangent space projection for a vector $H$ in the ambient space at $X\in\mathcal{M}$ is given by $\Proj_{X}\left(H\right):=H - \frac{\left<X,H\right>X}{\mu^2}$. The exponential map is given by $\Exp_{X}\left(tV\right)=\cos \left(\left\|V\right\|t\right) X + \sin \left(\left\|V\right\|t\right) \frac{\mu V}{\left\|V\right\|}$ for $V\in\calT_{X}\calM$ and $t\in\mathbb{R}$. 
% The Riemannian gradient is related to the Euclidean gradient via $\grad f\left(X\right) = \Proj_{X}\left(\nabla f\left(X\right)\right)$ where $\nabla f\left(X\right)$ is a vector in the ambient space. Also, the Riemannian Hessian seen as a linear operator applied to the tangent vector is related to the Euclidean Hessian via $\Hess f\left(X\right)\left[U\right] = \Proj_{X}\left(\nabla^{2}f\left(X\right)\left[U\right]\right) - \left<X,\nabla f\left(X\right)\right>U$ where $\nabla^{2}f\left(X\right)\left[U\right]$ is a vector in the ambient space.

% For a matrix $X=\left[x_{ij}\right] \in \mathbb{R}^{n \times m}$, we have $E\left[\left\|X\right\|^{2}\right] = E\left[\sum_{i}\sum_{j}x_{ij}^2\right] = nm E\left[x_{ij}^{2}\right]$, showing that $\left\|X\right\| \propto \sqrt{nm}$ for a randomly sampled $X$. Note that $E\left[x^{2}\right] = \frac{1}{3}$ for $x \sim \mathcal{U}_{\left[0,1\right]}$. The Frobenius norm of a matrix informally measures the variance (or variability) of its elements. Therefore, the matrices residing on a sphere have similar variance of elements.

\subsection{Deep Learning Formulation with Layerwise Normalisation} \label{SubSec:DLForm}
A machine learning problem is to train a function $f\left(x;w\right) \in \mathbb{R}^{d_{L}}$ of input $x\in\mathbb{R}^{d_{0}}$ with parameter vector $w$ by minimising an objective function $\calL\left(w\right) := \frac{1}{\left|\mathsf{D}\right|}\sum_{\left(x,y\right)\in \mathsf{D}} l\left(f\left(x;w\right),y\right)$ where $\mathsf{D}$ and $l$ denote the given data and convex loss function, respectively. Let us consider a fully-connected neural network model consisting of $L$ layers of function composition that can be represented as
\begin{equation} \label{Eq:FCNN}
    f\left(x;w\right):= \underbrace{W_{L}}_{\textrm{layer} L} \circ \underbrace{\rho \circ W_{L-1}}_{\textrm{layer} L-1} \circ \cdots \circ \underbrace{\rho \circ W_{1}}_{\textrm{layer} 1}\left(x\right)
\end{equation}
where $W_{i}\in\calM_{i}:=\mathbb{S}_{\mu_{i}}^{d_{i-1}d_{i}-1} \subset \mathbb{R}^{d_{i-1} \times d_{i}}$ is the $i$-th group of layerwise-normalised parameters in $w := \left(W_{1},\cdots,W_{L}\right)$, i.e., $\left\|W_{i}\right\|=\mu_{i}$, and $\rho$ represents the elementwise application of a nonlinear activation function.

A function $f:\mathcal{X} \rightarrow \mathcal{Y}$ between metric spaces $\left(\mathcal{X},d_{\mathcal{X}}\right)$ and $\left(\mathcal{Y},d_{\mathcal{Y}}\right)$ is called $K$-Lipschitz continuous if there exists a real nonnegative constant $K$ such that $d_{\mathcal{Y}}\left(f\left(x_{1}\right), f\left(x_{2}\right)\right) \leq K d_{\mathcal{X}}\left(x_{1},x_{2}\right)$. Given $K_{f}$-Lipschitz continuous $f: \mathcal{Y}\rightarrow \mathcal{Z}$ and $K_{g}$-Lipschitz continuous $g:\mathcal{X} \rightarrow \mathcal{Y}$, the composite function $f \circ g: \mathcal{X}\rightarrow \mathcal{Z}$ is also $K_{f}K_{g}$-Lipschitz continuous since $d_{\mathcal{Z}}\left(f\left(g\left(x_{1}\right)\right), f\left(g\left(x_{2}\right)\right)\right) \leq K_{f}d_{\mathcal{Y}}\left(g\left(x_{1}\right), g\left(x_{2}\right)\right) \leq K_{f}K_{g}d_{\mathcal{X}}\left(x_{1},x_{2}\right)$. It is obvious that the composition of $K_{f_{i}}$-Lipschitz continuous functions $f_{i}$ leads to a $\prod_{i} K_{f_{i}}$-Lipschitz continuous function. Therefore, if $\rho$ is $K_{\rho}$-Lipschitz continuous, the model in \eqref{Eq:FCNN} is $\prod_{i=1}^{L}\left\|W_{i}\right\|_{op}K_{\rho}^{L-1}$-Lipschitz continuous in $x$ with respect to the Euclidean distance. 

In this study, we will consider the nonlinear activation such that $\left|\rho\left(x\right)\right|\leq \left|x\right|$, $\left|\rho\left(x\right)-\rho\left(y\right)\right|\leq \left|x-y\right|$, and $\left|\rho'\left(x\right)\right|\leq 1$ for $x\in\mathbb{R}$. Under this prescription, $\left\|\rho\left(v\right)\right\|\leq \left\|v\right\|$ for $v\in\mathbb{R}^{n}$, and therefore,
\begin{equation} \label{Eq:Bound_FCNN}
    \left\|f\left(x;w\right)\right\| \leq \prod_{i=1}^{L}\left\|W_{i}\right\|_{op} \left\|x\right\| % \leq \prod_{l=1}^{L}\left\|W_{l}\right\| \left\|x\right\| = \prod_{l=1}^{L}\mu_{l} \left\|x\right\|
\end{equation}

\section{Automatic Optimisation on Combined Matrix Spheres} \label{Sec:ARGD}
\subsection{Initialisation}
The initialisation step for a normalised neural network specifies the values of $\mu_{i}$. Since $\left\|W_{i}\right\|_{op} \leq \mu_{i}$ for $i=1,\cdots,L$ and the model prediction always satisfies Eq. \eqref{Eq:Bound_FCNN}, it is necessary to scale the data during the preprocessing step so that the dataset conforms to the gain bound given in Eq. \eqref{Eq:Bound_FCNN}. Algorithm \ref{Algo:init} summarises the initialisation procedure for using normalised neural networks.

\begin{algorithm2e}[h!t]
    \caption{Spectral Initialisation} \label{Algo:init}
    \KwData{unnormalised $\mathsf{D}$, $\mu_{i}$ for $i=1,\cdots,L$}
    \KwResult{normalised $\mathsf{D}$, $W_{i}$ for $i=1,\cdots,L$}

    \For{$i \in 1:L$}{
        Randomly initialise $W_{i}$

        $W_{i} \gets \mu_{i} \frac{W_{i}}{\left\| W_{i} \right\|}$
    }

    $Y_{\max} \gets \max\limits_{y\in \mathsf{D}_{\text{train}}}\left\|y\right\|$

    \For{$\left(x,y\right) \in \mathsf{D} = \mathsf{D}_{\text{train}}\cup\mathsf{D}_{\text{test}}$}{
        $x \gets \frac{x}{\left\|x\right\|}$

        $y \gets \prod_{i=1}^{L}\left\|W_{i}\right\|_{op} \frac{y}{Y_{\max}}$
    }
\end{algorithm2e}

\subsection{Update}
The parameter $w$ of \eqref{Eq:FCNN} lies on a product manifold $\calM := \calM_{1} \times \cdots \times \calM_{L}$ of spheres $M_{i}=\mathbb{S}_{\mu_{i}}^{d_{i-1}d_{i}-1}$ for $i=1,\cdots,L$. The Euclidean gradient-based optimisation algorithms are inconsistent with the geometry of $\mathcal{M}$. Each iteration of the Euclidean Newton method has the additive update structure given by $w' = w + v$ where $v$ solves the following subproblem:
\begin{equation} \label{Eq:EucN}
    \min_{v\in\mathbb{R}^{\dim\left(w\right)}} m_{w}^{E}\left(v\right) = \calL\left(w\right) + \nabla_{\gamma} \calL\left(w\right) v + \frac{1}{2}v^{T}\nabla^{2}\calL\left(w\right)v
\end{equation}
On the other hand, the Riemannian optimisation algorithms account for the Riemannian metric structure. Each iteration of the Riemannian Newton method has the update structure given by $w' = \Exp_{w}\left(v\right)$ where $v$ solves the following subproblem:
\begin{equation} \label{Eq:RieN}
    \min_{v\in \calT_{w}\calM} m_{w}^{R}\left(v\right) = \calL\left(w\right) + \left<\grad \calL\left(w\right), v\right>_{w} + \frac{1}{2}\left<\Hess \calL \left(w\right)\left[v\right],v\right>_{w}
\end{equation}
However, the advantage of the Riemannian gradient descent as compared to the projected Euclidean gradient descent is not clear, and the Newton methods are too expensive for deep learning.

This study presents a method that considers the non-Euclidean geometry of $\calM$ while avoiding the expensive computation of the Riemannian Newton methods. Let us focus on the subproblem for each step of an iterative optimisation algorithm. Consider the differentiable curve $\gamma \subset \mathcal{M}$ that begins at $w$ and ends at $w'$. If $\gamma$ can be parameterised with respect to $t$ in $\left[0,\tau\right]$, then the update curve can be represented as $\gamma\left(t\right)$ where $\gamma\left(0\right) = w$ and $\gamma\left(\tau\right) = w'$. Since the parameter vector can be rearranged and grouped into matrices by layers, the update curve $\gamma$ can be represented as a collection of curves $\Gamma_{i}$ defined on $\mathcal{M}_{i}$ for $i=1,\cdots,L$, i.e., $\gamma\left(t\right) = \left(\Gamma_{1}\left(t\right),\cdots,\Gamma_{L}\left(t\right)\right)$. Correspondingly, we can parameterise the layer update curve as follows:
\begin{equation} \label{Eq:gamma_defn}
    \Gamma_{i}\left(t\right):=\Exp_{W_{i}}\left(t V_{i}\right) = W_{i} \cos t  + \mu_{i}V_{i} \sin t, \quad t \in \left[0,\tau\right]
\end{equation}
for $i=1,\cdots, L$ where $V_{i} \in \calT_{W_{i}}\calM_{i}$ denotes the update direction vector satisfying $\left\|V_{i}\right\|=1$ and $\left<W_{i},V_{i}\right>=0$. It should be noted that $\Gamma_{i}\left(t\right) = \Gamma_{i}\left(t+2N\pi\right)$ for any $N\in\mathbb{Z}$ and thus $\gamma\left(0\right)=\gamma\left(2\pi\right)$.

Both the update direction $V_{i}$ and the stepsize parameter $\tau$ should be designed to complete an optimisation algorithm. If $V_{i}$ is given for $i = 1,\cdots,L$, the remaining task is to find $\tau$. In this case, automatic determination of $\tau$ will provide an optimisation algorithm that does not require manual learning-rate tuning. In the followings, we develop methods to determine $\tau$ assuming that %the update direction is specified as
\begin{equation} \label{Eq:V_l}
    V_{i} = -\left.\frac{\Proj_{W_{i}}\left(\nabla_{\Gamma_{i}}\calL\left(\gamma\left(t\right)\right)\right)}{\left\| \Proj_{W_{i}}\left(\nabla_{\Gamma_{i}}\calL\left(\gamma\left(t\right)\right)\right)\right\|}\right|_{t=0}\qquad \text{for~} i = 1,\cdots,L
\end{equation} 
We will use the notation $\left.\nabla_{\Gamma_{i}}\calL\left(\gamma\left(t\right)\right)\right|_{t=0} = \nabla_{\Gamma_{i}}\calL\left(\gamma\left(0\right)\right)=\nabla_{\Gamma_{i}}\calL\left(w\right)$ for simplicity.

\subsubsection{Method 1. Hessian-Free Automatic Differentiation}
A direct method is to utilise AD technique. The objective function evaluated on $\gamma\left(t\right)$ becomes a function $\bar{\calL}\left(t\right):=\calL\left(\gamma\left(t\right)\right)$ of single variable $t$. Unlike the subproblems for Newton methods shown in Eqs. \eqref{Eq:EucN} and \eqref{Eq:RieN} that are based on the Taylor expansion taken at the level of $w$, the subproblem of the proposed method takes the Taylor expansion at the level of $t$ which amounts to computation of the directional derivatives of $\calL$ on $\calM$. The $r$-th order polynomial approximation is given by
\begin{equation} \label{Eq:Taylor_t}
    \bar{\calL}_{r}\left(t\right) := \sum_{k=0}^{r}\frac{\bar{\calL}^{\left(k\right)}\left(0\right)}{k!}t^{k}
\end{equation}
where the higher-order coefficients for $k > 2$ can be obtained by using Taylor-mode AD \cite{Tan_2022, Tan_2023}. The first coefficient can be computed as shown below using the information of $\nabla_{\Gamma_{i}}\calL\left(w\right)$ which is also used to determine $V_{i}$ in \eqref{Eq:V_l}.
\begin{equation} \label{Eq:dL_1}
    \begin{aligned}
        \bar{\calL}^{\left(1\right)}\left(0\right) &= \left<\nabla_{\gamma}\calL\left(\gamma\left(0\right)\right), \dot{\gamma}\left(0\right)\right>=\sum_{i=1}^{L}\left<\nabla_{\Gamma_{i}}\calL\left(\gamma\left(0\right)\right),\dot{\Gamma}_{i}\left(0\right)\right> = \sum_{i=1}^{L}\left<\nabla_{\Gamma_{i}}\calL\left(w\right),\mu_{i}V_{i}\right> \\
        &= -\sum_{i=1}^{L}\sqrt{\left\|\mu_{i}\nabla_{\Gamma_{i}}\calL\left(w\right)\right\|^{2} - \left<\nabla_{\Gamma_{i}}\calL\left(w\right), W_{i}\right>^{2}}
    \end{aligned}
\end{equation}
The second coefficient can be computed without explicit construction of Hessian by using forward-over-reverse AD \cite{Pearlmutter_1994,Martens_2010}.

The approach proceeds by finding 
% \begin{equation} \label{Eq:tau_Taylor_t}
$\tau =\underset{t\in\left[0,2\pi\right]}{\arg\min}\bar{\calL}_{r}\left(t\right)$.
% \end{equation}
One may consider reducing the range of $t$ to a trust region $\left[0, \epsilon\right]$ and solve the minimisation problem with $r=2$. Typical choice of the trust region is to set $\epsilon = \frac{\pi}{6}$ considering the validity of the small-angle approximation of trigonometric functions up to the second-order. The extreme point of $\bar{\calL}_{2}\left(t\right)$ is located at $t_{*}:=-\frac{\bar{\calL}^{\left(1\right)}\left(0\right)}{\bar{\calL}^{\left(2\right)}\left(0\right)}$. In this case, $\tau$ can be chosen among $\left\{0, \epsilon, t_{*}\right\}$ by comparing the objective value at each point in principle. By assuming high enough accuracy of the second-order approximation within the trust region, the stepsize can be determined without additional objective function evaluation. One such procedure can be summarised as in Algorithm \ref{Algo:tau_Taylor_t}. %Note that taking $\tau = 0$ is the decision to stop optimisation at that point.

% \begin{algorithm}[ht!]
%     \caption{Stepsize Determination Logic} \label{Algo:tau_Taylor_t}
%     \KwData{$\calL\left(\gamma\left(0\right)\right)=\calL\left(w\right)$, $\bar{\calL}^{\left(1\right)}\left(0\right)=-\sum_{l=1}^{L}\sqrt{\left\|\nabla_{\Gamma_{l}}\calL\left(w\right)\right\| - \left<W_{l},\nabla_{\Gamma_{l}}\calL\left(w\right)\right>^{2}}$, $\bar{\calL}^{\left(2\right)}\left(0\right)$}
%     \KwResult{$\tau$}
%     ~

%     $t_{*} \gets -\frac{\bar{\calL}^{\left(1\right)}\left(0\right)}{\bar{\calL}^{\left(2\right)}\left(0\right)}$

%     \eIf{$0\leq t_{*} \leq\epsilon$}{
%         \eIf{$\bar{\calL}^{\left(2\right)}\left(0\right) \geq 0$}{
%             $\tau \gets t_{*}$
%         }{
%             \eIf{$t_{*} \leq \frac{\epsilon}{2}$}{
%                 $\tau \gets \epsilon$
%             }{
%                 $\tau \gets 0$
%             }
%         }
%     }{
%         \eIf{$t_{*} > \epsilon$}{
%             \eIf{$\bar{\calL}^{\left(2\right)}\left(0\right) \geq 0$}{
%                 $\tau \gets \epsilon$
%             }{
%                 $\tau \gets 0$
%             }
%         }{
%             \eIf{$\bar{\calL}^{\left(2\right)}\left(0\right) \geq 0$}{
%                 $\tau \gets 0$
%             }{
%                 $\tau \gets \epsilon$
%             }
%         }
%     }
% \end{algorithm}

\begin{algorithm2e}[h!t]
    \caption{Stepsize Determination Logic} \label{Algo:tau_Taylor_t}
    \KwData{$\epsilon$, $\bar{\calL}^{\left(1\right)}\left(0\right)$, $\bar{\calL}^{\left(2\right)}\left(0\right)$}
    \KwResult{$\tau$}

    \eIf{$\bar{\calL}^{\left(2\right)}\left(0\right) = 0$}{
        \eIf{$\bar{\calL}^{\left(1\right)}\left(0\right) = 0$}{
            $\tau \gets 0$
        }{
            $\tau \gets \epsilon$
        }
    }{
        $t_{*} \gets -\frac{\bar{\calL}^{\left(1\right)}\left(0\right)}{\bar{\calL}^{\left(2\right)}\left(0\right)}$

        \eIf{$0\leq t_{*} \leq\epsilon$}{
            $\tau \gets t_{*}$
        }{
            $\tau \gets \epsilon$
        }
    }
\end{algorithm2e}

\subsubsection{Method 2. Majorisation-Minimisation}
The MM framework has been suggested as a general meta-theory for developing optimisation rules \cite{Lange_2016, Streeter_2023a,Bernstein_2023a, Bernstein_2023b, Landeros_2023}. Notably, \cite{Bernstein_2023a} noticed that the MM approach fits well with training of deep neural networks where high-dimensional optimisation should be performed without relying on expensive matrix inversion. To instantiate this approach to design optimisers, the key challenge lies in finding the majorant that can be derived and manipulated analytically.

% \subsubsection{Method 2-A. Expansion in Function}
The objective function evaluated along a update curve is a nested composition of functions that can be written as $\calL\left( \gamma\left(t\right) \right) = \frac{1}{\left|\mathsf{D}\right|}\sum_{\left(x,y\right)\in \mathsf{D}} l\left(f\left(x;\gamma\left(t\right)\right),y\right)$. The chain rule of differentiation indicates that the perturbation introduced into a variable affects the function value through the product of sensitivities for the intermediate variables between the function output and the perturbed variable. At each iteration, the movement along the update curve entails the following perturbations.
\begin{equation} \label{Eq:perts_defn}
    \begin{aligned}
        \Delta \gamma &:= \gamma\left(t\right) - \gamma\left(0\right) &\quad (\text{parameter perturbation})\\
        \Delta f_{x} &:= f\left(x;\gamma\left(t\right)\right) - f\left(x;\gamma\left(0\right)\right)&\quad  (\text{functional perturbation})
    \end{aligned}
\end{equation}

Consider the perturbation in the outermost variable $f$ in the definition of $l$. Expanding the loss function in $f$ leads to 
\begin{equation} \label{Eq:L_pert_f}
    % \begin{aligned}
        \resizebox{0.93\hsize}{!}{$\displaystyle
    l\left(f\left(x;\gamma\left(t\right)\right),y\right) = l\left(f\left(x;\gamma\left(0\right)\right),y\right) + \left<\nabla_{f}l\left(f\left(x;\gamma\left(0\right)\right),y\right), \Delta f_{x}\right> + \frac{1}{2}  \Delta f_{x}^{T} \nabla_{f}^{2}l\left(f\left(x;\gamma\left(0\right)\right),y\right) \Delta f_{x}+ \cdots
    $}
    % \end{aligned}
\end{equation}
For the squared $\ell^{2}$-norm $l\left(f\left(x;\gamma\left(t\right)\right),y\right) = \frac{1}{2}\left\| f\left(x;\gamma\left(t\right)\right)-y\right\|^{2}$, the higher-order derivatives in Eq. \eqref{Eq:L_pert_f} vanish as follows:
\begin{equation} \label{Eq:L_pert_f_l2}
    l\left(f\left(x;\gamma\left(t\right)\right),y\right) = l\left(f\left(x;\gamma\left(0\right)\right),y\right) + \left<f\left(x;\gamma\left(0\right)\right)-y, \Delta f_{x}\right> + \frac{1}{2} \left\| \Delta f_{x} \right\|^{2}
\end{equation}

To proceed further in the majorisation process, let us introduce a simplifying assumption for the first-order perturbation. 
\begin{assum} \label{Assum:first_pert}
    The first-order expansion of $l$ in $f$ is approximately equal to that in $\gamma$ at $t=0$, i.e.,
    \begin{equation} \label{Eq:first_pert}
        % \begin{aligned}
        \resizebox{0.93\hsize}{!}{$\displaystyle
        \left<\nabla_{f}l\left(f\left(x;\gamma\left(0\right)\right),y\right), \Delta f_{x}\right> \approx \left<\nabla_{\gamma}l\left(f\left(x;\gamma\left(0\right)\right),y\right), \Delta \gamma\right> = \sum_{i=1}^{L} \left<\nabla_{\Gamma_{i}} l\left(f\left(x;\gamma\left(0\right)\right),y\right), \Delta \Gamma_{i}\left(t\right)\right>
        $}
        % \end{aligned}
    \end{equation}
    where
    \begin{equation} \label{Eq:Delta_Gamma_l}
        % \begin{aligned}
            \Delta \Gamma_{i}\left(t\right) := \Gamma_{i}\left(t\right) - \Gamma_{i}\left(0\right) 
            = W_{i}\left(\cos t - 1\right) + \mu_{i}V_{i}\sin t
        % \end{aligned}
    \end{equation}
    according to the prescription of update curve $\gamma\left(t\right)$ given in Eq. \eqref{Eq:gamma_defn}.    
\end{assum}
The physical meaning of Assumption \ref{Assum:first_pert} is to neglect the contribution of the terms that are higher-order in $\Delta \gamma$ from the inner product $\nabla_{f} l$ and $\Delta f_{x}$. %Note that 
% \begin{equation} \label{Eq:norm_Delta_Gamma_l}
%     \left\|\Delta \Gamma_{i}\left(t\right)\right\| = 2\mu_{i}\sin \frac{t}{2} 
% \end{equation}
Substituting Eq. \eqref{Eq:first_pert} into Eq. \eqref{Eq:L_pert_f_l2} gives
\begin{equation}\label{Eq:L_pert_f_l2_t}
    l\left(f\left(x;\gamma\left(t\right)\right),y\right) = l\left(f\left(x;\gamma\left(0\right)\right),y\right) + \sum_{i=1}^{L} \left<\nabla_{\Gamma_{i}} l\left(f\left(x;\gamma\left(0\right)\right),y\right) , \Delta \Gamma_{i}\left(t\right)\right> + \frac{1}{2} \left\| \Delta f_{x} \right\|^{2}
\end{equation}
Summing up Eq. \eqref{Eq:L_pert_f_l2_t} over $\mathsf{D}$ and averaging gives
\begin{equation} \label{Eq:L_pert_f_l2_t_sum}
    % \begin{aligned}
        \calL\left(\gamma\left(t\right)\right) =  \calL\left(w\right) + \sum_{i=1}^{L} \left<\nabla_{\Gamma_{i}}\calL\left(w\right), \Delta\Gamma_{i}\left(t\right)\right>  +  \frac{1}{2\left|\mathsf{D}\right|}\sum_{\left(x,y\right)\in \mathsf{D}} \left\| \Delta f_{x} \right\|^{2}
    % \end{aligned}
\end{equation}

For descent, we want $ \calL\left(\gamma\left(t\right)\right) \leq \calL\left(w\right)$. The last term in Eq. \eqref{Eq:L_pert_f_l2_t_sum} represents the magnitude of the functional perturbation. Let us find an upper bound for this term to find a majorant for $\calL\left(\gamma\left(t\right)\right)$. For convenience, let us recursively define the latent states as follows:
\begin{equation} \label{Eq:latent_defn}
    \begin{aligned}
        h_{0}\left(x\right) &:= x\\
        h_{l}\left(x;s\right) &:= \rho\left(\Gamma_{l}\left(s\right)h_{l-1}\left(x;s\right)\right) ~\text{for}~ l=1,\cdots,L-1\\
        f\left(x;\gamma\left(s\right)\right) &= \Gamma_{L}\left(s\right)h_{L-1}\left(x;s\right)
    \end{aligned}
\end{equation}
with $\rho\left(\cdot\right)$ being $1$-Lipschitz with respect to $\ell^{2}$-norm as explained in Sec. \ref{SubSec:DLForm}. Then, perturbation at the $l$-th layer latent state can be bounded as
\begin{equation} \label{Eq:latent_pert}
    \begin{aligned}
        \left\| \Delta h_{l}\left(t\right)\right\| %&:=\left\| h_{l}\left(x;t\right) - h_{l}\left(x;0\right)\right\| \\
        &= \left\| \rho\left(\Gamma_{l}\left(t\right)h_{l-1}\left(x;t\right)\right) - \rho\left(\Gamma_{l}\left(0\right)h_{l-1}\left(x;0\right)\right)\right\| \\
        &\leq \left\| \Gamma_{l}\left(t\right)h_{l-1}\left(x;t\right) - \Gamma_{l}\left(0\right)h_{l-1}\left(x;0\right)\right\|\\
        &= \left\| \Gamma_{l}\left(t\right)\left[h_{l-1}\left(x;t\right)-h_{l-1}\left(x;0\right)\right] +\left[\Gamma_{l}\left(t\right)- \Gamma_{l}\left(0\right)\right]h_{l-1}\left(x;0\right)\right\|\\
        &\leq \left\| \Gamma_{l}\left(t\right)\left[h_{l-1}\left(x;t\right)-h_{l-1}\left(x;0\right)\right]\right\| +\left\|\left[\Gamma_{l}\left(t\right)- \Gamma_{l}\left(0\right)\right]h_{l-1}\left(x;0\right)\right\|\\
        &\leq %\left\| \Gamma_{l}\left(t\right)\right\|_{op}\left\|h_{l-1}\left(x;t\right)-h_{l-1}\left(x;0\right)\right\| +\left\|\Gamma_{l}\left(t\right)- \Gamma_{l}\left(0\right)\right\|_{op}\left\|h_{l-1}\left(x;0\right)\right\|\\
        %&:=
        \left\| \Gamma_{l}\left(t\right)\right\|_{op}\left\|\Delta h_{l-1}\left(t\right)\right\| +\left\|\Delta \Gamma_{l}\left(t\right)\right\|_{op}\left\|h_{l-1}\left(x;0\right)\right\|
    \end{aligned}
\end{equation}
where 
\begin{equation} \label{Eq:Delta_defn}
    % \begin{aligned}
        \Delta h_{l}\left(t\right) := h_{l}\left(x;t\right) - h_{l}\left(x;0\right) \\
    % \end{aligned}
\end{equation}
Since the Lipschitzness of the activation function leads to $\left\|h_{l-1}\left(x;s\right)\right\|\leq \prod_{i=1}^{l-1} \left\|\Gamma_{i}\left(s\right)\right\|_{op}\left\|x\right\|$, unrolling the recursive structure of the upper bound in Eq. \eqref{Eq:latent_pert} until reaching $l=0$ yields
\begin{equation} \label{Eq:Delta_h_l}
    % \begin{aligned}
        \left\| \Delta h_{l}\left(t\right)\right\| \leq \prod_{i=1}^{l}\left\| \Gamma_{i}\left(t\right)\right\|_{op}\underbrace{\left\|\Delta h_{0}\left(t\right)\right\|}_{=0} + \sum_{i=1}^{l} \prod_{j=i+1}^{l} \left\| \Gamma_{j}\left(t\right)\right\|_{op} \left\|\Delta \Gamma_{i}\left(t\right)\right\|_{op}\left\|h_{i-1}\left(x;0\right)\right\|\\
        % &\leq  \sum_{i=1}^{l} \prod_{j=i+1}^{l} \left\| \Gamma_{j}\left(t\right)\right\|_{op} \left\|\Delta \Gamma_{i}\left(t\right)\right\|_{op}\prod_{j=1}^{i-1} \left\|\Gamma_{j}\left(0\right)\right\|_{op}\left\|x\right\|
    % \end{aligned}
\end{equation}
and thus 
\begin{equation} \label{Eq:Delta_f_x}
    \left\| \Delta f_{x}\right\| \leq \sum_{i=1}^{L} \prod_{j=i+1}^{L} \left\| \Gamma_{j}\left(t\right)\right\|_{op} \left\|\Delta \Gamma_{i}\left(t\right)\right\|_{op}\prod_{j=1}^{i-1} \left\|\Gamma_{j}\left(0\right)\right\|_{op}\left\|x\right\|
\end{equation}
Using the triangle inequality $\left\| \Gamma_{j}\left(t\right)\right\|_{op} = \left\| \Gamma_{j}\left(0\right) + \Delta \Gamma_{j}\left(t\right)\right\|_{op} \leq \left\| \Gamma_{j}\left(0\right)\right\|_{op} + \left\| \Delta \Gamma_{j}\left(t\right)\right\|_{op}$ in Eq. \eqref{Eq:Delta_f_x} leads to
\begin{equation} \label{Eq:Delta_f_x_trust}
    \begin{aligned}
        &\left\| \Delta f_{x}\right\|\leq \sum_{i=1}^{L} \prod_{j=i+1}^{L} \left(\left\| \Gamma_{j}\left(0\right)\right\|_{op} + \left\| \Delta \Gamma_{j}\left(t\right)\right\|_{op} \right) \left\|\Delta \Gamma_{i}\left(t\right)\right\|_{op}\prod_{j=1}^{i-1} \left\|\Gamma_{j}\left(0\right)\right\|_{op}\left\|x\right\|\\
        &= \left\{\left\|\Delta \Gamma_{L}\left(t\right)\right\|_{op}\left\|\Gamma_{L-1}\left(0\right)\right\|_{op} \cdots \left\|\Gamma_{1}\left(0\right)\right\|_{op} \right.\\
        &\quad + \left(\left\|\Gamma_{L}\left(0\right)\right\|_{op} + \left\|\Delta \Gamma_{L}\left(t\right)\right\|_{op}\right) \left\|\Delta \Gamma_{L-1}\left(t\right)\right\|_{op} \cdots \left\|\Gamma_{1}\left(0\right)\right\|_{op}
        % &\quad + \left(\left\|\Gamma_{L}\left(0\right)\right\|_{op} + \left\|\Delta \Gamma_{L}\left(t\right)\right\|_{op}\right)  \left(\left\|\Gamma_{L-1}\left(0\right)\right\|_{op} + \left\|\Delta \Gamma_{L-1}\left(t\right)\right\|_{op}\right) \cdots \left\|\Gamma_{1}\left(0\right)\right\|_{op}\\
        \quad + \cdots \\
        &\quad + \left. \left(\left\|\Gamma_{L}\left(0\right)\right\|_{op} + \left\|\Delta \Gamma_{L}\left(t\right)\right\|_{op}\right)  \cdots \left(\left\|\Gamma_{2}\left(0\right)\right\|_{op} + \left\|\Delta \Gamma_{2}\left(t\right)\right\|_{op}\right) \left\|\Delta \Gamma_{1}\left(t\right)\right\|_{op} \right\} \left\|x\right\|\\
        &= \left[\prod_{i=1}^{L}\left(\left\| \Gamma_{i}\left(0\right)\right\|_{op} + \left\| \Delta \Gamma_{i}\left(t\right)\right\|_{op}\right) - \prod_{i=1}^{L} \left\| \Gamma_{i}\left(0\right)\right\|_{op} \right] \left\|x\right\|
    \end{aligned}   
\end{equation}
The upper bound on the right-hand-side of Eq. \eqref{Eq:Delta_f_x_trust} is the same as the deep relative trust proposed in \cite{Bernstein_2020}. %Since $\left\|\cdot\right\|_{op} \leq \left\| \cdot \right\|$, Eq. \eqref{Eq:Delta_f_x_trust} can be further bounded as
% \begin{equation} \label{Eq:Delta_f_x_t}
%     \begin{aligned}
%         \left\| \Delta f_{x} \right\| &\leq   \left[\prod_{i=1}^{L}\left(\left\| \Gamma_{i}\left(0\right)\right\| + \left\| \Delta \Gamma_{i}\left(t\right)\right\|\right) - \prod_{i=1}^{L} \left\| \Gamma_{i}\left(0\right)\right\| \right] \left\|x\right\|\\
%         &=\prod_{i=1}^{L}\mu_{i}\left[ \left(1 + 2\sin \frac{t}{2} \right)^{L} - 1 \right] \left\|x\right\|
%     \end{aligned}
% \end{equation}
Finally, substituting Eq. \eqref{Eq:Delta_f_x_trust} into Eq. \eqref{Eq:L_pert_f_l2_t_sum} and using Eqs. \eqref{Eq:Delta_Gamma_l} with \eqref{Eq:dL_1} gives the majorant as
\begin{equation} \label{Eq:L_pert_major}
    % \begin{aligned}
        \calL\left(\gamma\left(t\right)\right) \leq  \calL\left(w\right) +  \alpha \left(\cos t - 1\right) - \beta \sin t  + \frac{Q}{2} \left( P_{1}\left(t\right)-P_{2} \right)^{2}
    % \end{aligned}
\end{equation}
where $Q := \frac{1}{\left|\mathsf{D}\right|}\sum_{x\in \mathsf{D}} \left\| x \right\|^{2}$ and
\begin{equation} \label{Eq:alpha_beta_defn}
    \begin{aligned}
        \alpha &:= \sum_{i=1}^{L} \left<\nabla_{\Gamma_{i}}\calL\left(w\right), W_{i} \right> \qquad\qquad
        \beta := \sum_{i=1}^{L}\sqrt{\left\|\mu_{i}\nabla_{\Gamma_{i}}\calL\left(w\right)\right\|^{2} - \left<\nabla_{\Gamma_{i}}\calL\left(w\right), W_{i}\right>^{2}}\\
        P_{1}\left(t\right) &:= \prod_{i=1}^{L}\left(\left\| W_{i}\right\|_{op} + \left\| W_{i}\left(\cos t - 1\right) + \mu_{i}V_{i}\sin t\right\|_{op}\right)  \qquad\qquad
        P_{2} := \prod_{i=1}^{L}\left\|W_{i}\right\|_{op}       
    \end{aligned}
\end{equation}
The constant $Q$ is computed at the pre-processing stage and the other constants are computed at each iteration of the optimisation process.

Given the majorisation in Eq. \eqref{Eq:L_pert_major}, the MM procedure proceeds by minimising the RHS of Eq. \eqref{Eq:L_pert_major} to find $\tau \geq 0$. Since $\tau$ is an angular quantity that lies in a finite interval $\left[0,\pi\right]$, the problem is to minimise a univariate function on a bounded interval. Linesearch algorithms such as Brent's method or golden section search can be adopted to solve the minimisation problem in Algorithm \ref{Algo:tau_Taylor_MM}. The linesearch iteration does not require re-evaluation of the objective function and its gradient. % Note that the linesearch of the majorant is more computationally efficient than the linesearch of the original objective function since the function evaluation cost is cheaper for the majorant.

\begin{algorithm2e}[ht!]
    \caption{Linesearch for Stepsize Determination with Sinusoidal Majorisation} \label{Algo:tau_Taylor_MM}
    \KwData{$\alpha$, $\beta$, $P_{1}\left(t\right)$, $P_{2}$, $Q$}
    \KwResult{$\tau$}
    ~
    $\tau \gets \underset{t\in\left[0,\pi\right]}{\arg\min}\left[ \alpha \left(\cos t - 1\right) - \beta \sin t  + \frac{Q}{2} \left( P_{1}\left(t\right)-P_{2} \right)^{2} \right]$
\end{algorithm2e}

\section{Experiment} \label{Sec:NumExp}
We perform numerical experiments to demonstrate the characteristics of the proposed algorithms. The dataset for a quadrotor in ground effect provided in \cite{Shi_2019} is used for the experiments. The $12$ input variables include altitude, velocity, quaternion, and RPM, and the $3$ output variables are the aerodynamic forces. The number of datapoints is $28001$; $80\%$ of the randomly shuffled dataset is used for training and the rest is used for testing. We consider the loss function given by $l\left(f\left(x;w\right),y\right) = \frac{1}{2}\left\| f\left(x;w\right)-y\right\|^{2}$ for each $\left(x,y\right)\in\mathsf{D}$ normalised by Algorithm \ref{Algo:init}. In all cases, mini-batching is not applied, uniform Xavier initialisation is used, the parameters are normalised so that $\mu_{i}=1$ for all layers, and $\rho\left(x\right)=\mathrm{ReLU}\left(x\right)$ is used. Training is repeated for each case using $40$ different random seeds. The experiments are performed on a Mac mini 2020 with Apple M1 CPU and 16GB RAM. Our code developed in \texttt{Julia} is available via \cite{Cho_2023}.

\subsection{Experiment 1. Learning Trajectories for Fixed Width/Depth}
The first experiment aims to compare learning trajectories of the AD-based method 1 and the MM-based method 2 for a fixed combination of network width and depth. The layer configuration is set to be the same as \cite{Shi_2019}, i.e., $\begin{bmatrix}d_{0} & d_{1} & d_{2} & d_{3} & d_{4}\end{bmatrix} = \begin{bmatrix}12 & 25 & 30 & 15 & 3\end{bmatrix}$. Table \ref{Table:Exp1} shows the root-mean-square (RMS) value of the prediction error for training and test datasets along with the training time in the mean $\pm 1\sigma$ format. Figure \ref{Fig_E1} shows training/test objectives and the automatically determined stepsize with the mean and $1\sigma$ interval over 40 different random seeds. Method 1 provides much faster convergence than method 2 while achieving better prediction accuracies with less iterations. The rapid convergence is observed especially in the initial few tens of iterations. It turns out that the stepsize determined by method 2 is about two orders of magnitude smaller than method 1, which is thought to be resulting from the convervative nature of the chosen majorisation. The stepsize trajectory is more smooth and regular in method 2. A remedy is to scale up the stepsize determined by method 2 with a constant factor, however this will introduce a new hyperparameter. The stepsize determined by method 1 does not show significant variations after the initial 50 iterations while showing an overall decreasing pattern without any manual scheduling.

\begin{figure}[h!btp]
    \floatconts
    {Fig_E1}
    {\vspace{-1em}\caption{Objective Function Evaluations for Training and Test Dataset with Automatic Stepsize}}
    {
        \subfigure[Method 1 (AD)]{
            \includegraphics[width=0.48\textwidth]{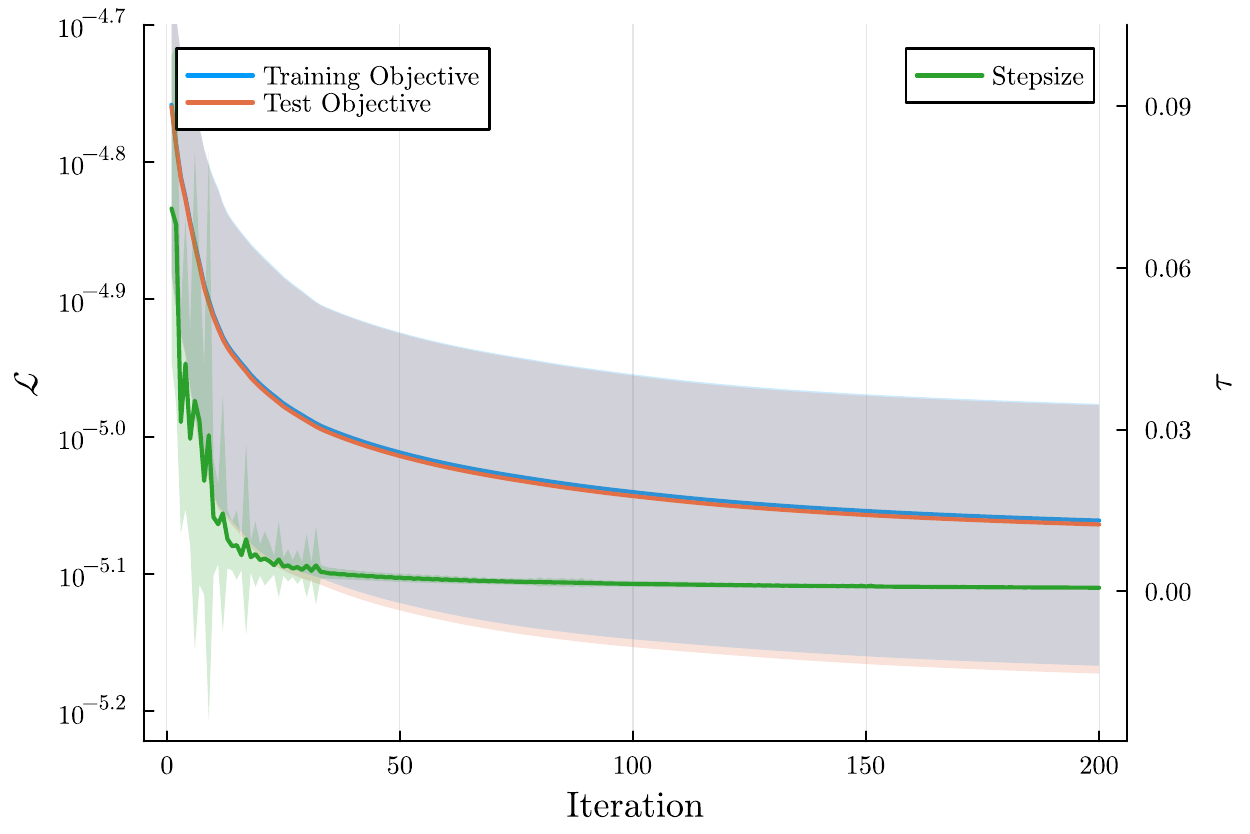}
        }
        \subfigure[Method 2 (MM)]{
            \includegraphics[width=0.48\textwidth]{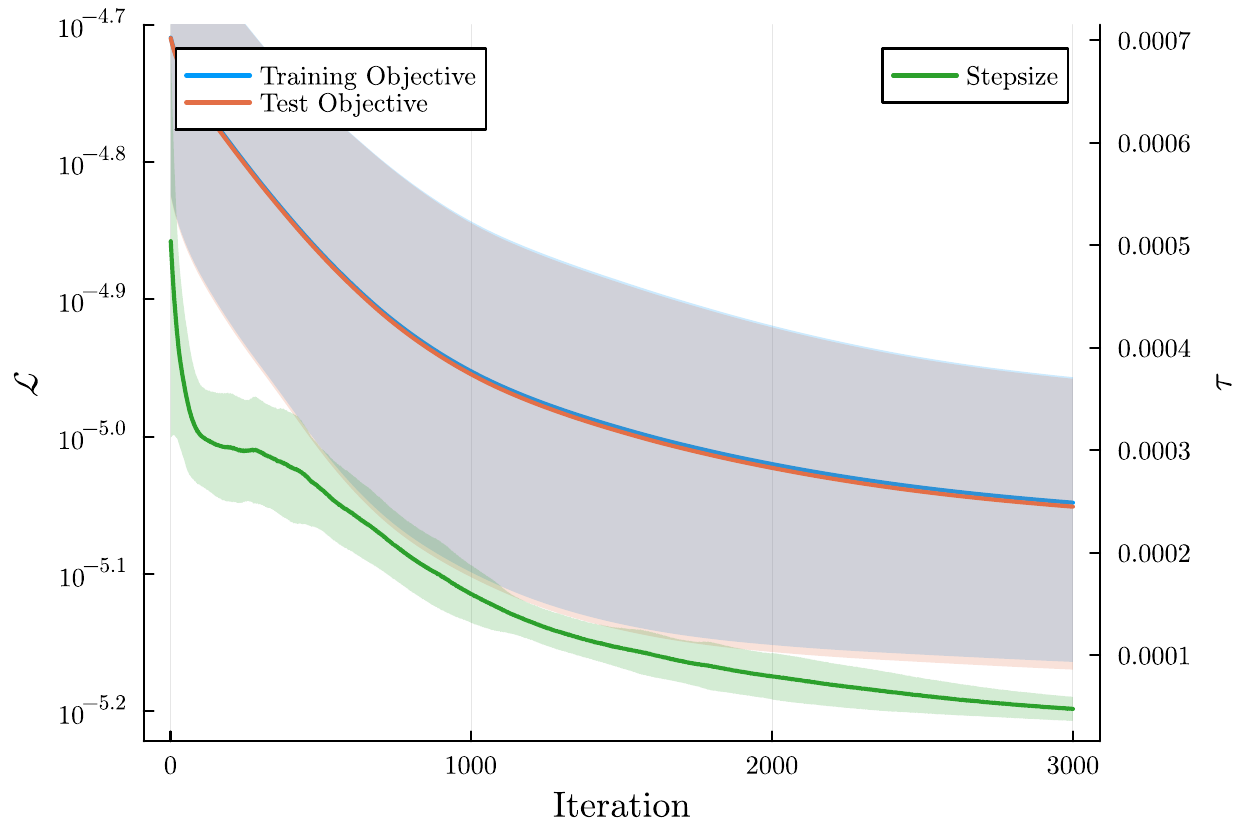}
        }
    }
\end{figure}

\begin{table}[!hbt]
    \begin{center}
        \begin{tabular}{ccccc}
            \hline
            Method  &   Training RMS Error [N]  &   Test RMS Error [N]  & Training Time [s]\\
            \hline
            1       &  $0.837 \pm 0.00658$    &   $0.834 \pm 0.0153$    &   $17.107 \pm 0.403$ / ~200 iterations\\
            2       &  $0.848 \pm 0.0138$     &   $0.845 \pm 0.0204$    &   $54.416 \pm 3.998$ / 3000 iterations\\
            \hline
        \end{tabular}
        \caption{Prediction Accuracy and Training Time for Experiment 1}
        \label{Table:Exp1}
    \end{center}
    \vspace{-1em}
\end{table}

\subsection{Experiment 2. Learning Results for Various Widths/Depths}
The second experiment aims to show the performance dependency of the AD-based method 1 with $200$ iterations on the network widths and depths. We consider a grid over widths $W \in 15:5:35$ and depths $L \in 4:2:12$ to set the layer configuration so that $d_{i}=W$ for $i=2,\cdots,L-1$ while $d_{0}=12$ and $d_{L}=3$. Figure \ref{Fig_E2} shows the RMS errors for training and test datasets and the training time with their mean values over 40 different random seeds for each width/depth combination. The heatmaps for the mean RMS errors display consistent trends. First, the proposed method achieves a similar level ($<1$N) of prediction accuracies for all layer configurations. Second, for each given depth, increasing width tends to yield smaller RMS errors for both training and test datasets. Third, for each given width, increasing depth tends to yield larger RMS errors. Fourth, the variation of RMS error is larger across different depths than across widths. Lastly, the present benchmark shows that the proposed method is particularly effective for training smaller networks in terms of achieved accuracies with respect to training time.
% Since the optimisation is stopped after a fixed number of iterations while the objective value descends at each iteration, it is expected that a target accuracy can be achieved with more iterations. 

\begin{figure}[h!btp]
    \floatconts
	    {Fig_E2}
	    {\vspace{-1em}\caption{Prediction Accuracy and Training Time for Various Layer Configurations}}
	    {
	        \subfigure[Training RMS Error {[N]}]{
	            \includegraphics[width=0.315\textwidth]{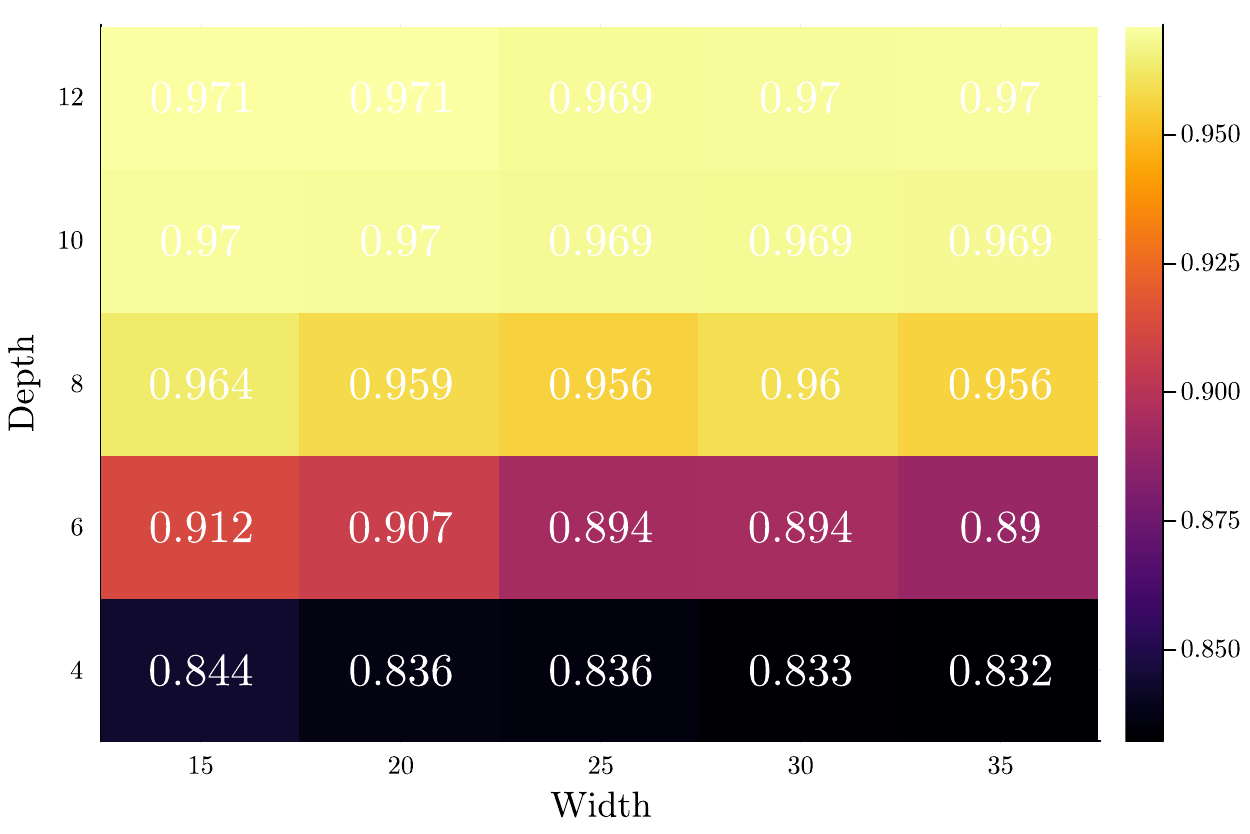}
	        }
	        \subfigure[Test RMS Error {[N]}]{
	            \includegraphics[width=0.315\textwidth]{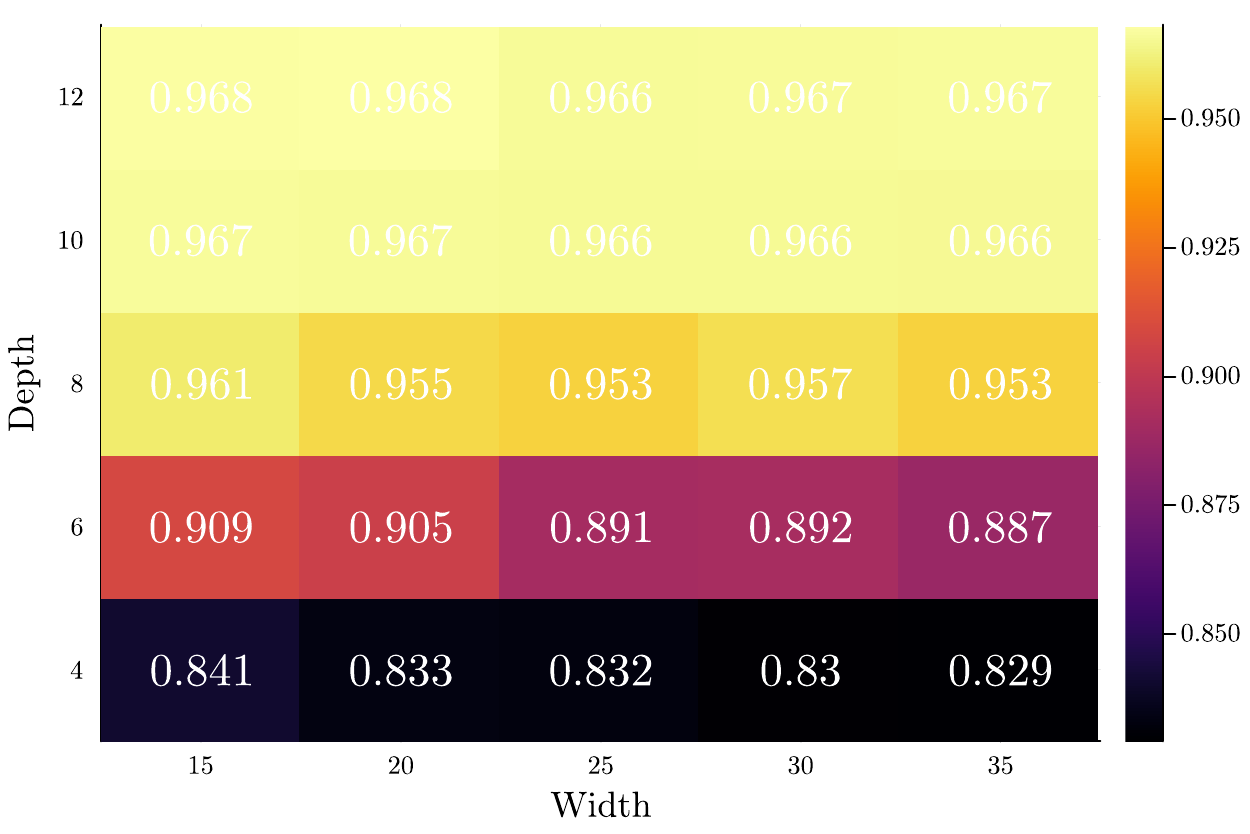}
	        }
	        \subfigure[Training Time {[s]}]{
	            \includegraphics[width=0.315\textwidth]{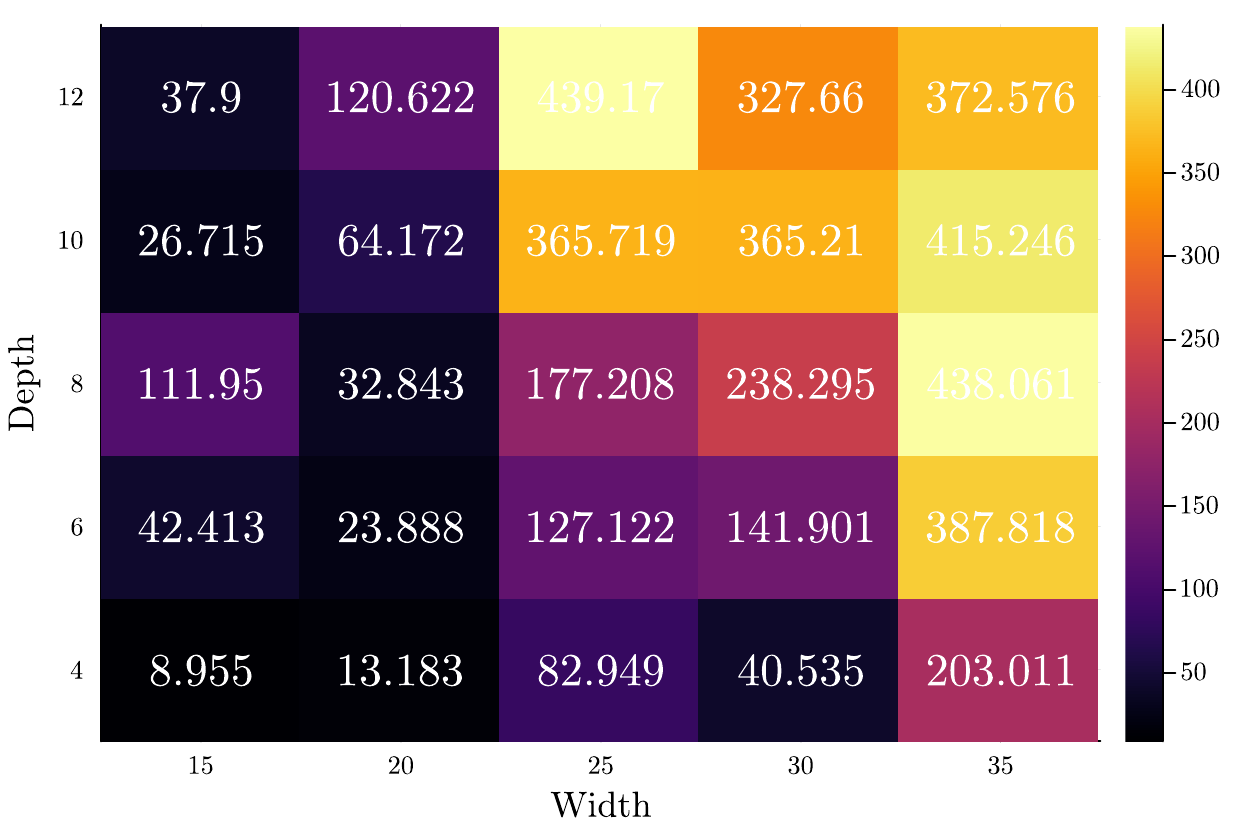}
	        }
	    }
%    \vspace{-1em}
\end{figure}

\section{Discussion} \label{Sec:Concls}
This study has proposed automatic optimisation methods for fully-connected neural networks with layerwise normalisation of parameters which specifies the Frobenius norm of the weights for each layer. Since the Frobenius norm can be computed more efficiently than the operator norm and is represented with a simple closed-form expression, evaluation and update of the neural network is easier with Frobenius normalisation than with spectral normalisation. For this type of normalised neural network, the proposed initialisation and update methods provide a complete neural network optimisation workflow that does not require manual tuning of the stepsize while explicitly accounting for the spherical motion of trainable parameters. Experiments on a dynamic model regression problem showed the effectiveness of the Hessian-free second-order stepsize determination method across a broad range of layer width/depth configurations. The proposed method achieved rapid convergence in the initial phase and consistent trends observed in the prediction accuracy without any manual learning rate scheduling or gradient clipping.

% Acknowledgments---Will not appear in anonymised version
% \acks{We thank ABC.}

\bibliography{ASGD}

\end{document}